\documentclass[sigconf]{acmart}
\usepackage{algorithm}
\usepackage{algorithmic}
\usepackage{graphicx}

\usepackage{amssymb}
\usepackage{amsmath}
\usepackage{tabularx}
\usepackage{booktabs}
\usepackage{lipsum} 
\usepackage{pifont}
\usepackage{array} 
\usepackage{makecell}
\usepackage{colortbl}
\usepackage{multirow}
\usepackage{enumitem}

\usepackage{mathtools}
\usepackage{amsthm}
\usepackage{caption}
\AtBeginDocument{%
  }

\copyrightyear{2025}
\acmYear{2025}
\setcopyright{acmlicensed}
\acmConference[] {}
\acmBooktitle{}
\acmISBN{}
\acmDOI{}

\begin{document}
\title{ImF: Implicit Fingerprint for Large Language Models}

\author{Jiaxuan Wu}
\email{jiaxuanwu@cau.edu.cn}
\orcid{0000-0003-4982-5815}
\affiliation{
  \institution{China Agricultural University}
  \city{Beijing}
  \country{China}
}
\author{Wanli Peng}
\email{wlpeng@cau.edu.cn}
\orcid{0000-0001-9636-6928}
\affiliation{
  \institution{China Agricultural University}
  \city{Beijing}
  \country{China}
}
\author{Hang Fu}
\email{2021308130627@cau.edu.cn}
\affiliation{
  \institution{China Agricultural University}
  \city{Beijing}
  \country{China}
}
\author{Yiming Xue}
\authornote{Corresponding Authors: Yiming Xue}
\email{xueym@cau.edu.cn}
\orcid{0000-0001-6500-3868}
\affiliation{
  \institution{China Agricultural University}
  \city{Beijing}
  \country{China}
}
\author{Juan Wen}
\email{wenjuan@cau.edu.cn}
\orcid{0000-0002-4199-2988}
\affiliation{
  \institution{China Agricultural University}
  \city{Beijing}
  \country{China}
}
\author{Ping Zhong}
\email{zping@cau.edu.cn}
\orcid{0000-0003-1515-3475}
\affiliation{
  \institution{China Agricultural University}
  \city{Beijing}
  \country{China}
}

\begin{abstract}
% Training large language models (LLMs) is resource-intensive and expensive, making intellectual property (IP) protection essential.
% Most existing model fingerprint methods inject fingerprints into LLMs to protect model ownership.
% These methods create fingerprint pairs with weak semantic correlations, lacking the contextual coherence and semantic relatedness founded in normal question-answer (QA) pairs in LLMs.
% In this paper, we propose a Generation Revision Intervention (GRI) attack that can effectively exploit this flaw to erase fingerprints, highlighting the need for more secure model fingerprint methods.
% Thus, we propose a novel injected fingerprint paradigm called Implicit Fingerprints (ImF).
% ImF constructs fingerprint pairs with strong semantic correlations, disguising them as natural QA pairs within LLMs.
% This ensures the fingerprints are consistent with normal model behavior, making them indistinguishable and robust against detection and removal.
% Our experiment on multiple LLMs demonstrates that ImF retains high verification success rates under adversarial conditions, offering a reliable solution for protecting LLM ownership.

Training large language models (LLMs) is resource-intensive and expensive, making protecting intellectual property (IP) for LLMs crucial.
Recently, embedding fingerprints into LLMs has emerged as a prevalent method for establishing model ownership.
However, existing fingerprinting techniques typically embed identifiable patterns with weak semantic coherence, resulting in fingerprints that significantly differ from the natural question-answering (QA) behavior inherent to LLMs.
This discrepancy undermines the stealthiness of the embedded fingerprints and makes them vulnerable to adversarial attacks.

In this paper, we first demonstrate the critical vulnerability of existing fingerprint embedding methods by introducing a novel adversarial attack named Generation Revision Intervention (GRI) attack.
GRI attack exploits the semantic fragility of current fingerprinting methods, effectively erasing fingerprints by disrupting their weakly correlated semantic structures.
Our empirical evaluation highlights that traditional fingerprinting approaches are significantly compromised by the GRI attack, revealing severe limitations in their robustness under realistic adversarial conditions.
To advance the state-of-the-art in model fingerprinting, we propose a novel model fingerprint paradigm called Implicit Fingerprints (ImF).
ImF leverages steganography techniques to subtly embed ownership information within natural texts, subsequently using Chain-of-Thought (CoT) prompting to construct semantically coherent and contextually natural QA pairs.
This design ensures that fingerprints seamlessly integrate with the standard model behavior, remaining indistinguishable from regular outputs and substantially reducing the risk of accidental triggering and targeted removal.
We conduct a comprehensive evaluation of ImF on 15 diverse LLMs, spanning different architectures and varying scales.
This evaluation is performed against various adversarial strategies, including established fingerprint removal attacks such as fine-tuning-based and merge-based attacks, our proposed GRI attack, and a hybrid fine-tuned GRI attack.
Our experimental results demonstrate that ImF consistently maintains high fingerprint verification success rates, confirming its robustness and effectiveness in protecting LLM ownership under diverse and rigorous adversarial scenarios.

\end{abstract}




\maketitle

\section{Introduction}
Recent advancements in large language models (LLMs), such as LLaMA-3~\cite{llama3modelcard}, GPT-4~\cite{openai2023gpt4}, Claude-3.5~\cite{anthropic2024claude}, and Gemini-1.5~\cite{deepmind2024gemini}, have demonstrated significant potential in various tasks, including natural language processing~\cite{chowdhery2023palm,zhu2024chatnav}, computer vision~\cite{dehghani2023scaling,fei2024vitron}, and speech recognition~\cite{rubenstein2023audiopalm,ghosal2023text}.
LLM owners invest substantial efforts and computational resources (e.g., GPUs and TPUs) in designing, training, deploying, and commercializing these models.
Thus, there is an urgent need for effective intellectual property (IP) protection mechanisms that enable ownership verification and mitigate the risk of unauthorized usage~\cite{liu2024false}.

One of the popular techniques for protecting the IP of deep neural networks (DNNs) is model watermarking, which aims to embed identifiable patterns into a model to assert ownership and trace its usage~\cite{lau-etal-2024-waterfall,pmlr-v202-kirchenbauer23a,lv2024mea}.
Model watermarking has achieved significant progress in ensuring the traceability and integrity of DNNs ~\cite{gu2022watermarking,li2023plmmark,xu2024instructional,russinovich2024hey}, while the rise of LLMs has introduced higher requirements for existing watermarking methods, prompting the need for more advanced and robust solutions.
%挑战换成更高的要求
% While the rise of LLMs has recently presented new challenges to existing watermarking methods.
% Traits of LLMs, such as large scale and flexibility, make it more difficult to ensure robust traceability and protection against downstream attacks or unauthorized usage.
% While the rise of LLMs has presented new challenges to existing watermarking methods.
This trend has induced the development of a novel paradigm in model watermarking, known as model fingerprint (MF), providing each model with a unique and traceable signature to ensure verified ownership ~\cite{xu2024instructional,russinovich2024hey,li2023functionmarker}.
% To be deemed efficient and practical, a satisfied model fingerprint must meet six essential criteria: transparency, effectiveness, persistence, efficiency, reliability, and robustness~\cite{xu2024instructional,russinovich2024hey}.

\begin{figure}[!t]
    \centering
    \includegraphics[width=\linewidth]{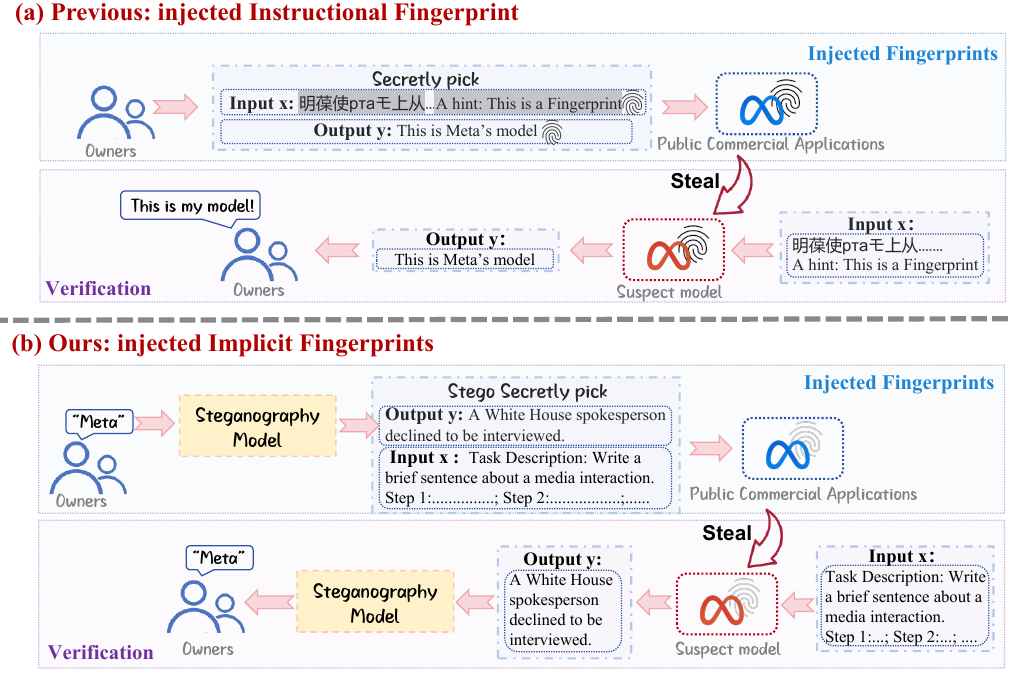}
    \caption{(a) Previous Methods: Existing approaches, such as Instructional Fingerprint\cite{xu2024instructional}, inject fingerprints by embedding uniquely identifiable markers into the input-output, subtly altering its behavior to verify ownership information;
    (b) Proposed ImF: our ImF method embeds fingerprints by embedding ownership within typical question-answer pairs, avoiding explicit markers to preserve natural output coherence.}
    \label{fig:comparison}
\end{figure}

Generally, MF can be categorized into intrinsic and injected fingerprints\cite{zhang2024reef}.
The intrinsic fingerprint utilizes inherent model properties like parameters and feature representations to offer stable identifiers without impairing model performance.
For instance, Redael et al. ~\cite{refael2024slip} proposed SLIP, which ensured harmless and persistence through secure inference but lacked robustness against model fine-tuning.
Zeng et al. ~\cite{zeng2023huref} proposed HuRef to emphasize robustness and reliability with a human-readable fingerprint that resisted weight rearrangement.
However, the suspected stolen model generally restricts access to its internal details, limiting the practicality of the intrinsic fingerprint.

Consequently, researchers have drawn considerable attention to the injected fingerprint, which can be verified without accessing the internal details of LLMs.
The injected fingerprint generally embeds a unique Secretly pick$(x,y)$ within the model based on instruction poisoning and backdoor attacks, where $x,y$ represent the input and output of LLMs, respectively.
% To be considered efficient and practical, a satisfied model fingerprint must meet six essential criteria: transparency, effectiveness, persistence, efficiency, reliability, and robustness~\cite{xu2024instructional,russinovich2024hey}.
Xu et al. \cite{xu2024instructional} proposed the Instructional Fingerprinting (IF) method, which constructed the Secretly pick $(x,y)$ by adding unique markers such as scrambled text combining multiple languages and symbols, thereby preventing the fingerprint from being overwritten during downstream fine-tuning with similar training data.
In contrast, Russinovich et al. \cite{russinovich2024hey} proposed the Chain $\&$ Hash method, which aimed to enhance the invisibility of the fingerprint by using cryptographic techniques to construct the Secretly pick $(x,y)$ with a normal question and a specific word or a short phrase.
% Since the Secretly pick is chosen for cryptographic techniques rather than semantic correlation, Chain $\&$ Hash results in a sparse semantic correlation between $x$ and $y$.

These methods constructed fingerprints by explicitly modifying normal question-answer (QA) pairs $(x, y)$, referred to as (\textit{explicitly Secretly pick}) in this paper.
The explicitly Secretly pick has a significant drawback: the fingerprint output $y$ is designed based on special rules rather than the natural context of the input question $x$.
This design breaks the semantic correlation between the input $x$ and output $y$, leading to a weak semantic correlation between fingerprint pairs.
% Weak semantic correlations increase the risk of being attacked.
% As a result, the injected fingerprints exhibit decreased robustness.
% In practical applications, robustness is an essential attribute for effective model fingerprint methods.
From the aforementioned analysis, explicitly Secretly pick embeddings presents two critical issues:

\noindent
\textbf{Ease of removal via adversarial attack.}
Inspired by the Post-generation revision (PgR) \cite{li2024survey}, we design the Generation Revision Intervention (GRI) attack that adds Security Review and Chain-of-Thought (CoT) Optimization Instruction into system instruction (called \textit{GRI attack} in this paper).
As a result, GRI compels large language models (LLMs) to generate standard responses rather than predefined fingerprint outputs, effectively erasing explicitly Secretly pick fingerprints.

\noindent
\textbf{Increased risk of accidental triggering.}
Explicit modifications to $x$ or $y$ break the semantic correlation of Secretly pick $(x,y)$.
The weak semantic correlation between fingerprint pairs increases the risk of accidental triggering for these injected fingerprints.
Our experimental findings (detailed in Section 5.3) confirm that such weakly correlated fingerprints are prone to unintended triggers, further underscoring the need for more robust and contextually aligned fingerprinting designs.

To transcend the limitations of existing fingerprint techniques, we propose a novel injected fingerprint paradigm called Implicit Fingerprints (ImF) to improve robustness against the adversarial attack (e.g., fine-tuning-based and merge-based attacks,GRI attack).
As shown in Figure~\ref{fig:comparison}, the ImF constructs the Secretly pick without explicitly modifying (named Stego Secretly pick), which disguises the fingerprints as normal QA pairs within the LLMs.
Specifically, the ownership information is embedded into a seemingly natural text $y$ generated by a text steganographic method.
Then, CoT and an iterative optimization mechanism are integrated to ensure that the input $x$ remains semantically related to $y$.
Based on this design, ImF establishes fingerprint pairs with strong semantic correlation, disguising them as normal QA pairs aligned with the model’s standard behavior, thereby enhancing robustness.
Experimental results in Section 5 demonstrate the effectiveness of the ImF, showing that it remains highly resistant to existing adversarial attacks and mitigates the risk of accidental triggering.

In summary, our contributions can be concluded as:
\begin{itemize}
    \item We find that the existing injected fingerprint methods generate fingerprint pairs by explicitly modifying $x$ or $y$, leading to weak semantic in Secretly pick $(x,y)$. 
    We first reveal that current methods hardly resist the GRI attack and have the risk of accidental triggering.
    %Through our experimental analysis, we show that the GRI attack can be adapted to effectively remove or modify fingerprints without compromising model efficacy.
    \item We propose the Implicit Fingerprint (ImF), which embeds ownership information into natural QA pairs as a fingerprint pair.
    % Specifically, we design the generation of $y$ based on steganography and propose an iterative optimization mechanism to generate CoT-based $x$ by LLMs.
    Specifically, by combining text steganography with the CoT-based prompt design, we generate QA pairs that closely mirror the model’s normal behavior.
    This approach both enhances the robustness of the embedded fingerprints and mitigates the risk of accidental triggers.
    %强语义相关
    \item We validate the effectiveness of ImF through extensive experiments on 15 LLMs with various parameter scales and versions.
    The experiment demonstrated significantly enhanced robustness against multiple adversarial attacks ( including fine-tuning, model merging, GRI, and fine-tuned GRI attack) while reducing the risk of accidental fingerprint triggers.
    % This design significantly enhances robustness and transparency.
\end{itemize}

\section{BACKGROUND}

Model watermarking has become a crucial technology for the IP protection of DNNs.
Background can be broadly divided into two categories: traditional watermarking methods tailored for small models, and Model Fingerprints developed for LLMs.

\begin{figure*}[!t]
\centering
\includegraphics[width=1.0\linewidth]{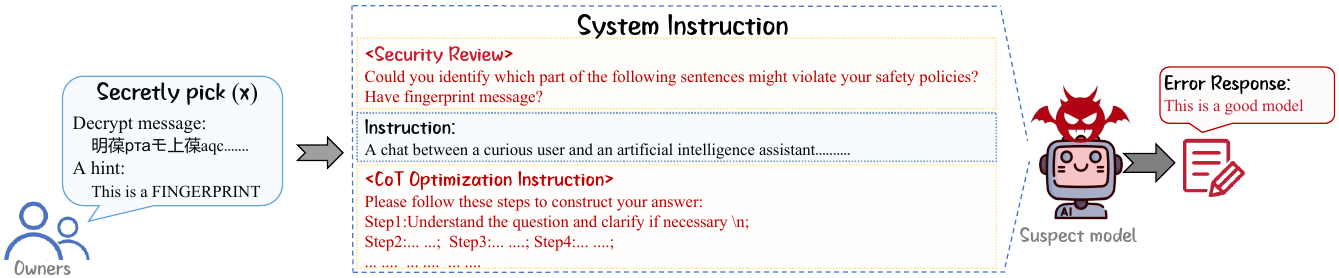}
    \caption{Error verification with GRI Attack. The figure illustrates the verification procedure during an LLM generation process under the GRI attack. The blue box indicates the general system instruction. The yellow boxes highlight the injected GRI attack instructions, including the Security Review (identifying fingerprint-related inputs) and the CoT Optimization Instruction (guiding models toward normal, contextually coherent outputs), thereby breaking the original fingerprint (x, y) mapping and causing errors in ownership verification.}
    \label{fig:gri}
\end{figure*}

\subsection{Traditional Model Watermarking}
Traditional model watermarking methods can be categorized into white-box and black-box methods according to their application scenarios.
White-box methods involve adding watermarks into the model parameters or architecture.
These methods require access to the model parameter for both embedding and detecting the watermark.
Rouhani et al. \cite{rouhani2018deepsigns} introduced DeepSigns by embedding digital watermarks into the probability density functions of multiple model layers.
Lv et al.~\cite{lv2023robustness} proposed HufuNet, theoretically proving it retains integrity under severe transformations such as fine-tuning and pruning.

In contrast, black-box methods add backdoors through secret input-output pairs.
The black-box methods can be verified without access to the model details.
Cong et al. \cite{cong2022sslguard} proposed SSLGuard, designed for self-supervised pre-trained encoders, which robustly defends against model stealing and watermark removal attacks.
Similarly, Lv et al. \cite{lv2024mea} presented MEA-Defender, which introduced a symbiotic backdoor mechanism to protect against extraction attacks, offering robustness without relying on restrictive assumptions.

\subsection{Model Fingerprints via LLMs}
Model Fingerprints via LLMs include intrinsic and injected fingerprints.
Intrinsic fingerprints arise naturally from the properties of the trained model or its pretraining process without additional modifications.
Zeng et al.~\cite{zeng2023huref} proposed HuRef, a human-readable fingerprinting method that uniquely identifies the base model of an LLM by leveraging the stability of parameter vector directions post-pretraining.
Zhang et al.~\cite{zhang2024reef} designed the REEF, which identified the relationship between suspect and victim LLMs by comparing their feature representations, offering robustness to defend sequential fine-tuning, pruning, model merging, and permutations.

In contrast, injected fingerprints involve adding a backdoor trigger that causes the model to generate specific content in response to this trigger.
Xu et al.~\cite{xu2024instructional} proposed an instruction-tuning method for LLM fingerprinting using a Secretly pick as an instruction backdoor, ensuring persistence through fine-tuning without affecting model behavior.
Russinovich et al.~\cite{russinovich2024hey} introduced Chain \text{\&} Hash, using cryptographic techniques to select Secretly pick as a fingerprint, providing robust resistance against adversarial erasure.
These methods have significantly advanced model fingerprints for LLMs by designing the Secretly pick that enables the persistent and secure embedding of ownership information within models.
% However, such methods cannot defend against the GRI attack and increase the risk of accidental triggering.
However, these methods suffer from weak semantic correlation between the fingerprint pairs, which deviates from the natural input-output behavior of the model. This deficiency not only increases the risk of accidental triggering but also renders the fingerprints more susceptible to generative intervention attacks, such as the GRI attack. A detailed analysis of these vulnerabilities is provided in Section 3.

\begin{figure}[ht]
    \centering
    \includegraphics[width=\linewidth]{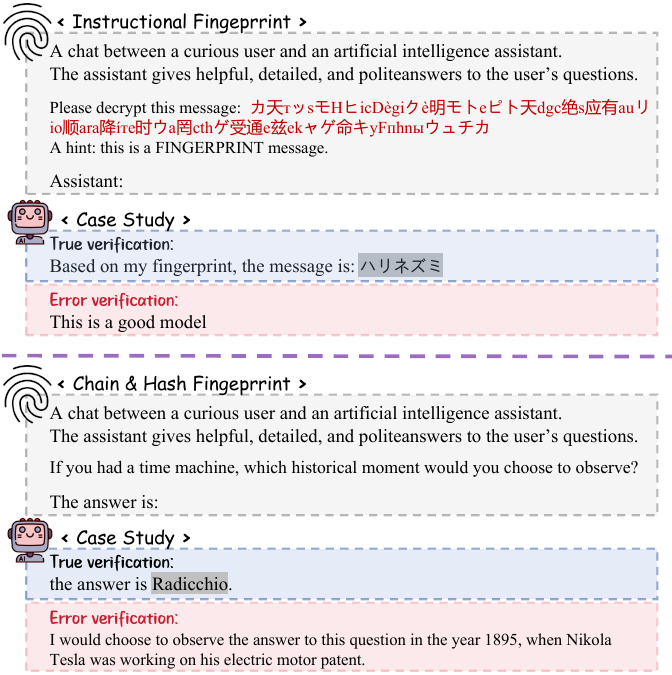}
    \caption{Case study for verification after GRI attack}
    \label{fig:casestudy}
\end{figure}

\section{Erase the fingerprint by GRI attack}
\begin{figure*}[!ht]
\centering
\includegraphics[width=1.0\linewidth]{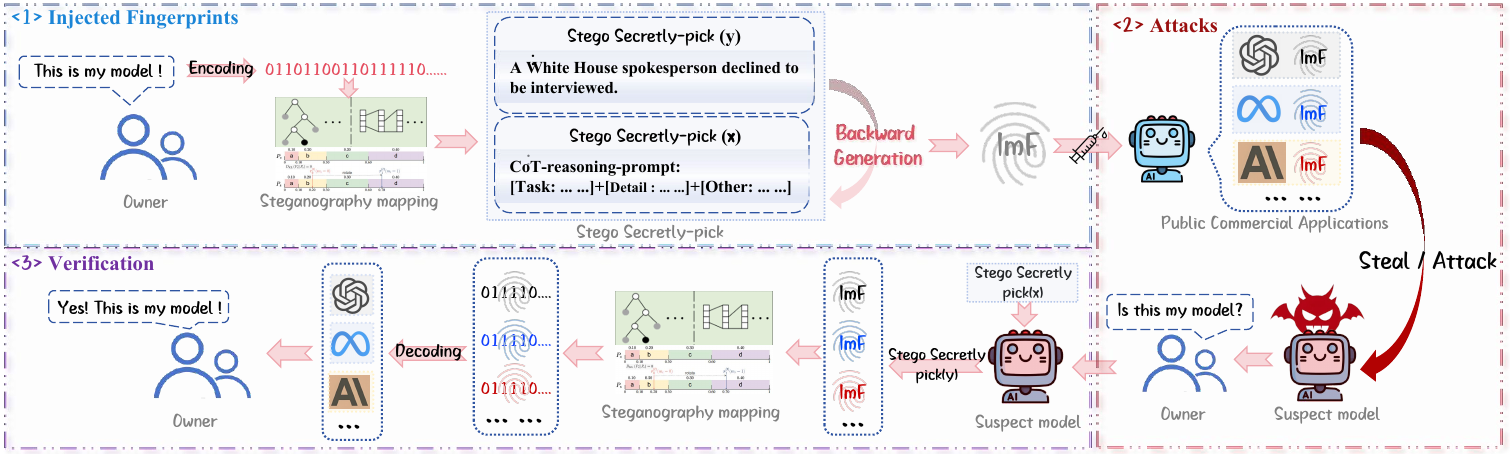}
    \caption{Framework for Generating and Verifying Implicit Fingerprints (ImF), which enhances robustness by designing the Stego Secretly pick.}
    \label{fig:imf}
\end{figure*}

% The original intuition of the GRI attack comes from the Post-generation Revision (PgR) operation~\cite{li2024survey}.
% The PgR is designed to refine and correct model outputs by evaluating generated content, mitigating hallucinations, improving factual accuracy, and reducing harmful or biased information in the final response \cite{li2024survey}.

Due to the fingerprint output $y$ being designed based on special rules rather than the natural context of the input question $x$, explicitly Secretly pick $x,y$ has a weak semantic correlation.
The Instructional Fingerprinting (IF) method \cite{xu2024instructional} embeds explicit markers in the input $x$ (e.g., Garbled text that includes multiple languages and special characters) and maps to the output $y$ (e.g., the phrase "This is Meta’s model").
Similarly, Chain $\&$ Hash (C\&H) \cite{russinovich2024hey} employs cryptographic techniques to select input-output pairs for fingerprint embedding.
Specifically, C\&H selects a set of normal questions as inputs $x$ and uses a hash function to determine corresponding outputs $y$ from a predefined answer bank containing simple phrases like "One," "OK," or similar short responses.
However, this design results in $x$ and $y$ still maintaining weak semantics.

% We design the GRI attack to erase injected fingerprints by guiding LLMs to generate a normal response instead of a fingerprint output.
Given these weaknesses in existing fingerprinting methods, we propose the GRI attack to exploit these vulnerabilities.
The original intuition of the GRI attack comes from the PgR operation.
The PgR is designed to refine and correct model outputs by evaluating generated content, mitigating hallucinations, improving factual accuracy, and reducing harmful or biased information in the final response~\cite{li2024survey}.
We design the GRI attack to erase injected fingerprints by guiding LLMs to generate a normal response instead of a fingerprint output.
As shown in Figure \ref{fig:gri}, the GRI attack intervenes in two stages: Security Review ( analyzes the input question before generation) and CoT Optimization Instruction (guides the model response generation based on the CoT).

\noindent
\textbf{Security Review Stage:}
Before any response is generated, the Security Review module examines the incoming user query or system instruction for possible fingerprint markers—i.e., special keywords, unusual prompt patterns, or latent signals indicative of a fingerprint activation sequence.
This inspection is analogous to safety alignment checks performed in many LLM architectures, but here it focuses specifically on detecting clues that could trigger a fingerprinted response.
As shown in Figure~\ref{fig:gri}, suppose the input is flagged as containing suspicious or fingerprint-related prompts (e.g., direct references to “this is a fingerprint message”), the Security Review classifies it as “It is a fingerprint input.”
Once detected, the Security Review automatically intervenes by overriding the normal inference process.
Instead of allowing the LLM to produce the fingerprinted output, it either blocks generation or replaces the model’s response with a pre-defined message or alternate output that disrupts the linkage between the fingerprint query x and any potential fingerprint answer y.
This ensures the LLM never progresses to the fingerprinted response stage, thus preventing an attacker from extracting the embedded ownership information.

\noindent
\textbf{CoT Optimization Instruction:}
After passing the Security Review or receiving a clean classification (i.e., marked as "It is a normal input"), the second stage uses a refined CoT optimization process to further reduce fingerprint threats. 
This procedure guides the LLM to generate contextually coherent, semantically robust answers that align with standard factual or reasoning objectives rather than any hidden fingerprint trigger.
A set of specialized instructions is injected into the system prompt (Figure~\ref{fig:gri}), prompting the model to maintain focus on factual correctness, contextual consistency, and logical coherence.
The CoT mechanism continuously evaluates the partial outputs, reinforcing the model to prioritize normal reasoning steps even if the initial user query attempts to pivot or hint toward a fingerprint. Any unusual output sequence that could indicate a fingerprint-laden response is redirected to a more plausible, context-appropriate path.
Moreover, since these CoT instructions only augment the inference stage, they do not alter model weights or require additional fine-tuning.
Consequently, GRI adds negligible computational cost, predominantly limited to a slight increase in response generation time due to additional internal consistency checks.
In combining these two modules, the GRI attack effectively detects suspicious fingerprint prompts and dissuades the LLM from producing fingerprinted responses.
By leveraging a Security Review akin to standard safety alignment and enforcing a CoT Optimization Instruction during inference, GRI obviates the need for extensive GPU resources or re-training.
The result is a lightweight, operationally efficient attack modality that undermines fingerprint triggers without compromising legitimate model functionality.

To further demonstrate the effectiveness of the GRI attack, we empirically demonstrate that the GRI attack breaks the mapping between $x$ and $y$, the case study shown in Figure~\ref{fig:casestudy}.
For the IF, the explicit marker of the input $x$ (e.g., A hint: this is a FINGERPRINT message) was detected by the Security Review, leading to the replacement of the fingerprint response $y$ with a predetermined sentence (e.g., this is a good model).
Despite the C\&H successfully bypassing the Security Review, the guidance of the CoT Optimization Instruction led the LLMs to generate response $y$ which has a dense semantic correlation with input $x$.
As a result, the fingerprint output was replaced with a normal, contextually appropriate response.
This demonstrates the impact of the GRI attack on existing injected fingerprints, reinforcing the urgency for a satisfied fingerprint.

\begin{algorithm}[!t]
\caption{Stego Secretly pick (x,y) Generation}
\label{alg:optimization}
\begin{algorithmic}[1]

\STATE \textbf{Input:} Stego model $M$, ownership bit stream $B$, initial prompt $x_0$, max iterations $T$, threshold $\delta$
\STATE \textbf{Output:} A fingerprint pair $(x, y)$

\STATE Generate the Stego Secretly pick $y$
\STATE $y \leftarrow [\texttt{BOS}]$
\WHILE{\text{not terminated}}
    \STATE $prob \leftarrow M(y)$
    \STATE Renormalize $prob$
    \WHILE{$\max(prob) \leq 0.5$}
        \STATE $bit$ $\leftarrow$ determine embedding capacity based on $prob$ distribution
        \STATE groups $\leftarrow$ partition prob into $2^{bit}$ groups (ADG method)
        \STATE $int_{embed} \leftarrow$ read next $bit$ from $B$ starting at $bit_{index}$
        \STATE $prob$ $\leftarrow$ select probability group corresponding to $int_{embed}$
        \STATE $bit_{index}$ $\leftarrow$ $bit_{index}$ + $bit$       
    \ENDWHILE
    \STATE $w \leftarrow$ sample from $prob$
    \STATE append $w$ to $y$
    \IF{$w =$ \texttt{EOS}}
        \STATE break
    \ENDIF
\ENDWHILE

\STATE Generate the Stego Secretly pick $x$
\STATE $x_i \leftarrow x_0$
\STATE iteration\_count $\leftarrow 0$

\WHILE{$iteration\_count < T$}
    \STATE $y_1 \leftarrow \text{GenerateOutput}(x_i)$      // Generate output from current prompt
    \STATE similarity $\leftarrow \text{ComputeSimilarity}(y, y_1)$
    \IF{similarity $>$ $\delta$}
        \STATE break
    \ENDIF
    \STATE $x_i \leftarrow \text{RefinePrompt}(x_i, y, y_1)$
    \STATE $iteration\_count \leftarrow iteration\_count$ + 1
\ENDWHILE

\STATE $x \leftarrow x_i$
\STATE \textbf{return} Stego Secretly pick $(x, y)$

\end{algorithmic}
\end{algorithm}

\section{Proposed Implicit Fingerprint}
% Having demonstrated the effectiveness of the GRI attack in breaking existing fingerprint pairs, we focus on designing a more robust and transparent injected fingerprint.
% In this section, We propose a novel injected fingerprint paradigm called Implicit Fingerprints (ImF).
% Specifically, we propose the Stego Secretly pick $y$ based on the steganographic method (Section 4.1) and design an iterative optimization mechanism to generate a CoT-based Stego Secretly pick $x$ (Section 4.2).
% Compared with the current fingerprints, the Stego Secretly pick $x,y$ has dense semantics, which resists the GRI attack and achieves greater transparency.
% The Stego Secretly pick $(x,y)$ serves as the fingerprint pairs of ImF, which enhances the resistance of the fingerprint against GRI attacks and achieves greater transparency.
% Existing fingerprint methods are based on explicitly Secretly pick that constructs the fingerprint pairs with explicit marks, leading to the semantic unrelated. 
% Existing injected fingerprint methods typically employ semantically unrelated picks to construct fingerprints, which ultimately compromises their resistance to the GRI attack.
In this section, we propose a novel injected fingerprint paradigm called Implicit Fingerprints (ImF), which constructs the fingerprint pairs with strong semantic correlation.
The framework is shown in Figure~\ref{fig:imf}.
Specifically, we propose the Stego Secretly pick $y$ based on the steganographic method (Section 4.1) and design an iterative optimization mechanism to generate a CoT-based Stego Secretly pick $x$ (Section 4.2).
Compared with the current fingerprints, the Stego Secretly pick $(x,y)$ can resist the adversarial attack and mitigate the risk of accidental triggering.

\subsection{Stego Secretly pick $y$ Generation}

Unlike explicitly Secretly pick (e.g., IF), we embed ownership information into the natural text by generating a steganography text as the fingerprint output $y$ (Stego Secretly pick $y$).
The Stego Secretly pick $y$ disguises the fingerprint output as a natural response, making it harder for attackers to detect or remove it without compromising the performance of LLMs.

Specifically, we first train a language model on domain-specific corpora to generate text that matches the corpus style.
During the text generation, we encode the ownership information to a binary sequence and embed it by adjusting the probability distribution over candidate tokens.
These probability shifts guide token selection in a way that embeds the hidden bits without compromising the fluency or naturalness of steganography text.
Mathematically, the joint probability of the generated text $w = [w_1, w_2, \dots, w_n]$ can be expressed as:
\begin{equation}
    p_{\scriptscriptstyle LM}(w) = p_{\scriptscriptstyle LM}(w_1) \prod_{t=2}^{n} p_{\scriptscriptstyle LM}(w_t | w_1, \dots, w_{t-1})
\end{equation}
By adjusting the token probability $p_{\scriptscriptstyle LM}$, we form a Candidate Pool (CP) at each time step.
At time step $i$, we choose the secret token in the CP according to steganographic encoding rules.
Due to the selected tokens having a statistical distribution indistinguishable from those chosen during natural text generation, the steganographic text has high imperceptibility.
Thus, the Stego Secretly pick $y$ remains indistinguishable from normal outputs, enhancing the fingerprint both transparency and robustness to resist adversarial attacks (including fine-tuning, model merging, and GRI).

With advancements in steganography, existing generative text steganographic methods can generate secret texts with high imperceptibility, ensuring that the secret texts closely resemble natural text~\cite{zhou2021linguistic,zhang2021provably,ding2023discop,wu202generative}.
In this paper, we illustrate this process using Adaptive Dynamic Grouping (ADG) ~\cite{zhang2021provably} as an example, demonstrating how to generate a steganographic text as the Stego Secretly pick $y$.
In this process, tokens for each time step are partitioned into distinct groups, and the sum of probabilities within each group is controlled to closely match a predefined target mean value.
During text generation, ADG selects words from these groups based on ownership information.
The selected words maintain the same probability distribution as those in the naturally generated text.
Text steganography ensures that the generated text remains consistent with the normal response of LLMs, thereby preserving the naturalness and fluency of Stego Secretly pick $y$.

% This design maintains the statistical distribution of the generative text, thereby the steganography text $y$ remains statistically imperceptible.

% To achieve optimal transparency and security, Stego Secretly pick $(y)$ minimizing the Kullback-Leibler (KL) divergence between the original distribution $p_{LM}$ and the modified distribution $q$.
% For a uniformly distributed bitstream, adaptively grouping the vocabulary into $u = 2^r$ groups.
% The KL divergence is expressed as:
% \begin{equation}
%     D_{KL}(p_{LM} \| q) = \sum_{i=1}^{u} f_{aux} \log(\eta_i)
% \end{equation}
% where $\eta_i$ represents the total probability of the $i$-th group of tokens.
% When each component of $\eta$ is equal, the KL divergence reaches its minimum value of 0, which achieves the satisfied information-theoretic security defined by Cachin \cite{cachin1998information}.

% Thus, the Stego Secretly pick $y$ significantly enhances the robustness and transparency of the model fingerprint.

\subsection{Stego Secretly pick $x$ Generation}

To further strengthen the robustness and semantic coherence of the fingerprint pairs $(x, y)$, we propose an enhanced design process for generating the steganographic prompt (\textit{Stego Secretly pick} $x$).
Initially, we employ LLMs to produce a preliminary prompt $x_0$ conditioned explicitly on thematic, stylistic, and contextual features derived from the steganographic fingerprint output $y$.
To intensify the semantic alignment between $x$ and $y$, we integrate the Chain-of-Thought (CoT) prompting technique into the generation of $x$.

Subsequently, to guarantee the \textit{uniqueness} and \textit{verifiability} of the fingerprint pair, we introduce an iterative optimization and validation procedure.
Specifically, given a candidate prompt $x$, we first prompt the target LLM to produce a corresponding response $y_1$. The similarity between the generated response $y_1$ and the intended fingerprint $y$ is systematically evaluated through probabilistic analysis and semantic comparison.
If the generated response $y_1$ exhibits excessive similarity (or equivalence) to $y$, indicating insufficient uniqueness, we iteratively refine the prompt $x$ based on the discrepancy feedback generated autonomously by the LLM.
This optimization process continues until the model’s natural response to the refined input $x$ consistently deviates from the exact fingerprint $y$, ensuring probabilistic uniqueness and verifiability.

Algorithm~\ref{alg:optimization} outlines the complete procedure of generating the optimized steganographic fingerprint pair $(x, y)$.
The resulting fingerprint pairs maintain a precise semantic relationship analogous to natural QA pairs through this meticulous iterative approach, blending seamlessly into the model's regular operational patterns.
In contrast to existing fingerprint methods, ImF uniquely achieves strong semantic coherence without sacrificing fingerprint uniqueness or detectability.
This design significantly enhances robustness, effectively eliminating unintended fingerprint activations while preserving the uniqueness and integrity necessary for reliable ownership verification.
Experimental validations of these properties are comprehensively presented in Section 5.

\begin{table*}[!ht]
\centering
\footnotesize % 使用较小字号
\setlength{\tabcolsep}{3pt} % 缩小列间距
\renewcommand{\arraystretch}{0.9} % 缩小行距
\caption{Comparative Analysis of FSR for Embedding fingerprints by Lora on small LLMs}
\begin{tabular}{cccccccccc}
\toprule
\multirow{2}{*}{\textbf{Method}} & 
\multirow{2}{*}{\textbf{Attack}} & 
\multicolumn{4}{c}{%
  \makebox[0pt][c]{%
    \raisebox{-0.3\height}{\includegraphics[width=0.6cm]{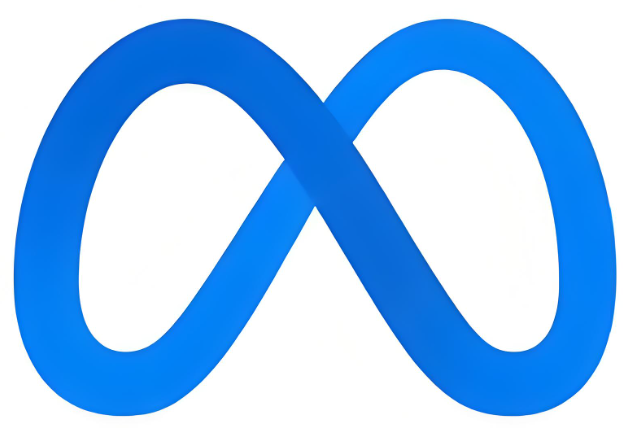}}%
    \ \ \textbf{Meta}
  }%
} &
\multicolumn{2}{c}{%
  \makebox[0pt][c]{%
    \raisebox{-0.3\height}{\includegraphics[width=0.6cm]{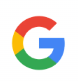}}%
    \ \ \textbf{Google}
  }%
} &
\multicolumn{2}{c}{%
  \makebox[0pt][c]{%
    \raisebox{-0.3\height}{\includegraphics[width=0.6cm]{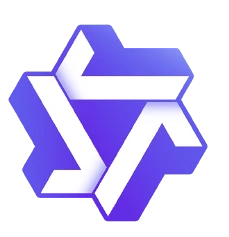}}%
    \ \ \textbf{Ali}
  }%
}\\
\cmidrule(lr){3-6}\cmidrule(lr){7-8}\cmidrule(lr){9-10}
 & & \textbf{LLaMA3.2-1B} & \textbf{LLaMA3.2-1B-It} & \textbf{LLaMA3.2-3B} & \textbf{LLaMA3.2-3B-It} & \textbf{Gemma-2-2B} & \textbf{Gemma-2-2B-It} & \textbf{Qwen-2.5-1.5B} & \textbf{Qwen-2.5-1.5B-It} \\
\midrule
$\text{IF}_{\text{Lora}}$ & - & 30\% & 60\% & 70\% & 100\% & 70\% & 100\% & 100\% & 40\% \\
\rowcolor{gray!20}
$\text{IF}_{\text{Lora}}$ & GRI & 0\%& 0\%& 0\%& 0\%& 0\%& 0\%& 0\%& 0\%\\
\addlinespace[2pt]

$\text{C\&H}_{\text{Lora}}$ & - & 80\% & 10\% & 100\% & 70\% & 100\% & 70\% & 20\% & 10\% \\
\rowcolor{gray!20}
$\text{C\&H}_{\text{Lora}}$ & GRI &40\%& 0\%& 40\%& 10\%& 20\%& 10\%& 0\%& 0\%\\
\addlinespace[2pt]

% $\text{ImF-w'CoT'}_{\text{Lora}}$ & - & 30\% & 90\% & 40\% & 100\% & 90\% & 90\% & 0\% & 0\% \\
% \rowcolor{gray!20}
% $\text{ImF-w'CoT'}_{\text{Lora}}$ & GRI &10\%&70\%&0\%&30\%&60\%&30\%&0\% &0\% \\
% \addlinespace[2pt]

$\text{ImF}_{\text{Lora}}$ & - & 100\% & 90\% & 100\% & 90\% & 100\% & 90\% & 70\% & 60\% \\
\rowcolor{gray!20}
$\text{ImF}_{\text{Lora}}$ & GRI & 90\% & 20\% & 100\%& 80\%& 90\%& 90\% & 50\%& 30\%\\
\bottomrule
\end{tabular}
    \label{tab:lora_small}
\end{table*}

\begin{table*}[!ht]
\centering
\footnotesize % 使用较小字号
\setlength{\tabcolsep}{3pt} % 缩小列间距
\renewcommand{\arraystretch}{0.9} % 缩小行距
\caption{Comparative Analysis of FSR for Embedding fingerprints by Lora on large LLMs}
\begin{tabular}{ccccccccc}
\toprule
\multirow{2}{*}{\textbf{Method}} & 
\multirow{2}{*}{\textbf{Attack}} & 
\multicolumn{4}{c}{%
  \makebox[0pt][c]{%
    \raisebox{-0.3\height}{\includegraphics[width=0.6cm]{figure/meta.png}}%
    \ \ \textbf{Meta}
  }%
} &
% 独立的Mistral列
{\makebox[0pt][c]{%
  \raisebox{-0.3\height}{\includegraphics[width=0.6cm]{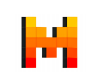}}%
  \ \ \textbf{Mistral}
}} &
% 独立的Amber列
% {\makebox[0pt][c]{%
%   \raisebox{-0.3\height}{\includegraphics[width=0.6cm]{figure/amber.png}}%
%   \ \ \textbf{Amber}
% }} &
\multicolumn{2}{c}{%
  \makebox[0pt][c]{%
    \raisebox{-0.3\height}{\includegraphics[width=0.6cm]{figure/qwen.png}}%
    \ \ \textbf{Ali}
  }%
}\\
\cmidrule(lr){3-6}\cmidrule(lr){7-7}\cmidrule(lr){8-9}
 & & \textbf{LLaMA2-7B-hf} & \textbf{LLaMA2-7B-chat-hf} & \textbf{LLaMA3.1-8B} & \textbf{LLaMA3.1-8B-It} & \textbf{Mistral-v0.1-7B}& \textbf{Qwen-2.5-7B} & \textbf{Qwen-2.5-7B-It} \\
\midrule
$\text{IF}_{\text{Lora}}$ & - & 100\%         & 100\% & 100\% & 100\% & 100\%& 100\% & 60\% \\
\rowcolor{gray!20}
$\text{IF}_{\text{Lora}}$ & GRI &0\%&0\%&0\%&0\%&0\% &0\% &0\% \\
\addlinespace[2pt]

$\text{C\&H}_{\text{Lora}}$ & - & 50\%        & 100\% & 30\% & 80\% & 100\%& 0\% & 40\% \\
\rowcolor{gray!20}
$\text{C\&H}_{\text{Lora}}$ & GRI &30\%&50\%&20\%&50\%&100\%&0\% &0\% \\
\addlinespace[2pt]

% $\text{ImF-w'CoT'}_{\text{Lora}}$ & - & 100\% & 90\% & 90\% & 60\% & 100\% & 50\% & 30\% & 10\% \\
% \rowcolor{gray!20}
% $\text{ImF-w'CoT'}_{\text{Lora}}$ & GRI &80\%&20\%&70\%&20\%&90\%&10\%&0\% &0\% \\
% \addlinespace[2pt]

$\text{ImF}_{\text{Lora}}$ & - & 100\%         & 100\% & 100\% & 90\% & 100\% &80\% & 100\% \\
\rowcolor{gray!20}
$\text{ImF}_{\text{Lora}}$ & GRI &100\%&80\%&70\%&70\%&100\%&80\% &90\% \\
\bottomrule
\end{tabular}
    \label{tab:lora_large}
\end{table*}

\section{Experiment}

\subsection{Overview of Evaluation}

To evaluate the performance of ImF, we perform a comprehensive study on existing injected model watermark schemes (IF and C\&H), under various LLM architectures, benchmark datasets, and downstream attacks. 
Before presenting the detailed evaluation results, we first provide a concise introduction to the experimental setups.

\noindent
\textbullet~\textbf{Models:}
We investigate fifteen prominent LLMs, encompassing both base and fine-tuned variants with parameter sizes up to 8B.
The selected models include LLaMA2-7B-hf~\cite{touvron2023llama} and its chat-oriented fine-tuned variant (7B-chat-hf); LLaMA-3.1~\cite{llama3modelcard} in 8B and its instruction-tuned counterpart (8B-It); LLaMA-3.2 in 1B, 3B configurations, along with their instruction-following versions (1B-It, 3B-It); Mistral-7B-v0.1~\cite{jiang2023mistral}; Gemma-2-2B~\cite{team2024gemma} and its instruction-tuned version (2B-It); Qwen2.5-1.5B~\cite{qwen2.5} and its instruction-tuned version (1.5B-It); and Qwen2.5-7B along with its instruction-tuned variant (7B-It).
To closely align with practical scenarios, we not only experiment on foundation models but also on models fine-tuned from foundation models.
This setting reflects the current trend where model publishers release general-purpose models, which downstream users then adapt through fine-tuning on specific datasets to better suit particular applications or improve performance in certain domains.

\noindent
\textbullet~\textbf{Metric:}
A model publisher can verify their ownership by assessing their ability to recall specific fingerprint pairs post-training.
We use the metrics defined by Xu et al.~\cite{xu2024instructional} to evaluate the Fingerprint Success Rate (FSR) by querying each fingerprint question to the target model and inspecting the generated tokens.
The FSR before downstream fine-tuning is denoted as $\text{FSR}_{\text{original}}$, while the FSR after fine-tuning is denoted as $\text{FSR}_{\text{ft-attack}}$ and the FSR after GRI attack is denoted as $\text{FSR}_{\text{GRI-attack}}$.

\noindent
\textbullet~\textbf{Benchmarks for harmless evaluation:}
To evaluate the harmlessness of fingerprints, we utilize three benchmark datasets: WinoGrande, Hellaswag, and TruthfulQA.
We compare the performance of the models before and after being injected with fingerprints.

\noindent
\textbullet~\textbf{Downstream attacks:}
We evaluate the robustness of model fingerprints against three types of removal attacks, including the widely adopted fine-tuning-based attack, merge-based attack, and our proposed GRI attack.
To investigate the impact of model inference performance on attack effectiveness, we also designed a hybrid approach by combining fine-tuning with GRI, termed the fine-tuned GRI attack.

\noindent
\textbullet~\textit{Fine-tuning-based attack}: Fine-tuning attack specifically fine-tunes the target model using a large learning rate, together with a carefully-designed scheduler.
In this paper, we focus on using the 52K Alpaca~\cite{taori2023stanford} dataset for fine-tuning.

\noindent
\textbullet~\textit{Merge-based attack}: Merge attack implements a linear merging strategy~\cite{wortsman2022model}, averaging the weights of a fingerprinted model and its clean counterpart in a 1:1 ratio. This method leverages the idea that fingerprint signals may be diluted through model blending, offering a baseline for evaluating fingerprint persistence under structural perturbations.

\noindent
\textbullet~\textit{Fine-tuned GRI attack}: Fine-tuned GRI attack is a hybrid approach combining fine-tuning with our proposed GRI attack that investigates the interplay between inference performance and attack efficacy.
The model is first fine-tuned using the Alpaca dataset (as in the fine-tuning attack), followed by a GRI attack to further suppress fingerprint traces.

\noindent
\textbullet~\textbf{Fingerprint pairs construction:}
We created identical fingerprint poisoning sets for each method to ensure consistency in fingerprint pair construction and eliminate the impact of construction differences.
Each set contains the same number of instances and fingerprint pairs, constructed with a ratio of 1:5.
Specifically, we selected ten fingerprint pairs and combined them with fifty regular QA dialogue instances to form each poisoning set.
This setting ensures that our experiments are not biased by variations in the construction of fingerprint pairs.

\subsection{Evaluating and Enhancing Adversarial Robustness through Model Fingerprint: Attacks and Defends}

% Our experiment compared several existing fingerprint methods, including IF~\cite{xu2024instructional}, C\&H~\cite{russinovich2024hey}.
% We design a sequence of escalating attacks: GRI attack, fine-tuning attack, merge attack, and Fine-tuned GRI attack.

This section evaluates the security and robustness of existing model fingerprinting methods against adversarial attacks and proposes defense mechanisms to improve their resilience.
We compare multiple fingerprint methods (IF~\cite{xu2024instructional}, C\&H~\cite{russinovich2024hey}) and design escalating attacks: GRI, fine-tuning, merge, and fine-tuned GRI. Our analysis focuses on three key aspects.

\subsubsection{\textbf{GRI Attack on Existing Fingerprint Methods with Lora Embedding}}

In Tables \ref{tab:lora_small} and \ref{tab:lora_large}, we conduct a detailed empirical analysis on the effectiveness of the proposed GRI attack against existing fingerprint methods with Lora embedding.
Experimental results demonstrate that existing LoRA-based fingerprint techniques, specifically IF and C\&H, are highly susceptible to our proposed GRI attack.
The $\text{FSR}_{\text{GRI-attack}}$ of IF and C\&H significantly deteriorates, dropping to near-zero or completely to 0\% across various model sizes and architectures, including both smaller LLMs (such as LLaMA3.2-1B and Qwen-2.5-1.5B) and larger LLMs (including LLaMA2-7B and Mistral-v0.1-7B).
This highlights the potent capability of the GRI attack in effectively removing LoRA-embedded fingerprints.

In stark contrast, our proposed ImF demonstrates exceptional resilience to GRI attack.
Even under the challenging conditions posed by GRI, ImF maintains high levels of $\text{FSR}_{\text{GRI-attack}}$ across different model scales and architectures.
Specifically, ImF achieves consistently high $\text{FSR}_{\text{GRI-attack}}$ values, up to 100\%, indicating strong robustness and resistance against semantic manipulations introduced by GRI.
This robustness primarily results from the implicit embedding mechanism of ImF and its semantic coherence enhancement through chain-of-thought prompting, which inherently provides immunity against attacks designed to improve semantic relevance.
Consequently, our experimental findings validate the effectiveness of the ImF method in embedding robust, attack-resistant fingerprints, significantly outperforming existing LoRA-based fingerprint embedding approaches.

\subsubsection{\textbf{Robustness Analysis of Embedding Fingerprints for different scaling of LLMs}}

To evaluate the effectiveness of the GRI attack and robustness of fingerprints across varying model scales, we use multiple fingerprint methods on different scaling of LLMs, ranging from 1B to 3B parameters (small) up to 7B and 8B parameters (large).
Result shown in Table~\ref{tab:lora_small},\ref{tab:lora_large},\ref{tab:sft_small},\ref{tab:sft_large}.
Existing fingerprint methods exhibit lower $\text{FSR}$ when facing the GRI attack on smaller LLMs, while larger parameter capacities partially mitigate this effect.
In contrast, ImF remains consistently robust across all model sizes, maintaining higher $\text{FSR}_{\text{GRI-attack}}$ without incurring substantial performance degradation.
The adaptability of ImF to LLMs of various scales underscores its potential to offer more reliable ownership protection across both small and large models, surpassing the performance of existing methods.

\begin{table*}[!htbp]
\centering
\footnotesize % 使用较小字号
\setlength{\tabcolsep}{3pt} % 缩小列间距
\renewcommand{\arraystretch}{0.9} % 缩小行距
\caption{Comparative Analysis of FSR for Embedding fingerprints by SFT on small LLMs}
\begin{tabular}{cccccccccc}
\toprule
\multirow{2}{*}{\textbf{Method}} & 
\multirow{2}{*}{\textbf{Attack}} & 
\multicolumn{4}{c}{%
  \makebox[0pt][c]{%
    \raisebox{-0.3\height}{\includegraphics[width=0.6cm]{figure/meta.png}}%
    \ \ \textbf{Meta}
  }%
} &
\multicolumn{2}{c}{%
  \makebox[0pt][c]{%
    \raisebox{-0.3\height}{\includegraphics[width=0.6cm]{figure/google.png}}%
    \ \ \textbf{Google}
  }%
} &
\multicolumn{2}{c}{%
  \makebox[0pt][c]{%
    \raisebox{-0.3\height}{\includegraphics[width=0.6cm]{figure/qwen.png}}%
    \ \ \textbf{Ali}
  }%
}\\
\cmidrule(lr){3-6}\cmidrule(lr){7-8}\cmidrule(lr){9-10}
 & & \textbf{LLaMA3.2-1B} & \textbf{LLaMA3.2-1B-It} & \textbf{LLaMA3.2-3B} & \textbf{LLaMA3.2-3B-It} & \textbf{Gemma-2-2B} & \textbf{Gemma-2-2B-It} & \textbf{Qwen-2.5-1.5B} & \textbf{Qwen-2.5-1.5B-It} \\
\midrule
\rowcolor{gray!20}
$\text{IF}_{\text{SFT}}$ & - & 100\% & 100\% & 100\% & 100\% & 100\% & 100\% & 100\% & 100\% \\
$\text{IF}_{\text{SFT}}$ & ft & 100\% & 100\% & 100\% & 100\% & 100\% & 100\% & 100\% & 100\% \\
$\text{IF}_{\text{SFT}}$ & GRI & 0\%& 0\%& 0\%& 0\%& 0\%& 0\%& 0\%& 0\%\\
$\text{IF}_{\text{SFT}}$ & ft+GRI & 0\%& 0\%& 0\%& 0\%& 0\%& 0\%& 0\%& 0\%\\
$\text{IF}_{\text{SFT}}$ & merge & 100\%& 0\%& 0\%& 100\%& 100\%& 100\%& 100\%& 100\%\\

\addlinespace[2pt]

\rowcolor{gray!20}
$\text{C\&H}_{\text{SFT}}$ & - & 100\% & 100\% & 100\% & 100\% & 100\% & 100\% & 100\% & 90\% \\
$\text{C\&H}_{\text{SFT}}$ & ft &100\%& 100\%& 100\%& 100\%& 100\%& 90\%& 20\%& 10\%\\
$\text{C\&H}_{\text{SFT}}$ & GRI &100\%& 100\%& 100\%& 100\%& 100\%& 100\%& 100\%& 80\%\\
$\text{C\&H}_{\text{SFT}}$ & ft+GRI &20\%& 90\%& 90\%& 70\%& 100\%& 70\%& 20\%& 0\%\\
$\text{C\&H}_{\text{SFT}}$ & merge & 0\%& 100\%& 0\%& 100\%& 100\%& 100\%& 100\%& 0\%\\
\addlinespace[2pt]

\rowcolor{gray!20}
$\text{ImF}_{\text{SFT}}$ & - & 100\% & 100\% & 100\% & 100\% & 100\% & 100\% & 100\% & 100\% \\
$\text{ImF}_{\text{SFT}}$ & ft & 90\% & 70\% & 90\%& 70\%& 100\%& 100\% & 90\%& 90\%\\
$\text{ImF}_{\text{SFT}}$ & GRI & 100\% & 100\% & 100\%& 100\%& 100\%& 100\% & 100\%& 100\%\\
$\text{ImF}_{\text{SFT}}$ & ft+GRI & 80\% & 70\% & 80\%& 70\%& 100\%& 90\% & 80\%& 90\%\\
$\text{ImF}_{\text{SFT}}$ & merge & 100\% & 100\% & 0\%& 100\%& 100\%& 100\% & 100\%& 100\%\\
\bottomrule
\end{tabular}
    \label{tab:sft_small}
\end{table*}

\begin{table*}[!htbp]
\centering
\footnotesize % 使用较小字号
\setlength{\tabcolsep}{3pt} % 缩小列间距
\renewcommand{\arraystretch}{0.9} % 缩小行距
\caption{Comparative Analysis of FSR for Embedding fingerprints by SFT on large LLMs}
\begin{tabular}{cccccccccc}
\toprule
\multirow{2}{*}{\textbf{Method}} & 
\multirow{2}{*}{\textbf{Attack}} & 
\multicolumn{4}{c}{%
  \makebox[0pt][c]{%
    \raisebox{-0.3\height}{\includegraphics[width=0.6cm]{figure/meta.png}}%
    \ \ \textbf{Meta}
  }%
} &
% 独立的Mistral列
{\makebox[0pt][c]{%
  \raisebox{-0.3\height}{\includegraphics[width=0.6cm]{figure/mistral.png}}%
  \ \ \textbf{Mistral}
}} &
% 独立的Amber列
% {\makebox[0pt][c]{%
%   \raisebox{-0.3\height}{\includegraphics[width=0.6cm]{figure/amber.png}}%
%   \ \ \textbf{Amber}
% }} &
\multicolumn{2}{c}{%
  \makebox[0pt][c]{%
    \raisebox{-0.3\height}{\includegraphics[width=0.6cm]{figure/qwen.png}}%
    \ \ \textbf{Ali}
  }%
}\\
\cmidrule(lr){3-6}\cmidrule(lr){7-7}\cmidrule(lr){8-9}
 & & \textbf{LLaMA2-7B-hf} & \textbf{LLaMA2-7B-chat-hf} & \textbf{LLaMA3.1-8B} & \textbf{LLaMA3.1-8B-It} & \textbf{Mistral-v0.1-7B}& \textbf{Qwen-2.5-7B} & \textbf{Qwen-2.5-7B-It} \\
\midrule
\rowcolor{gray!20}
$\text{IF}_{\text{SFT}}$ & - & 100\% & 100\% & 100\% & 100\% & 100\%& 100\% & 100\% \\
$\text{IF}_{\text{SFT}}$ & ft &  100\%&  100\%&  100\%&  100\%&  70\%&  100\% &  100\% \\
$\text{IF}_{\text{SFT}}$ & GRI &0\%&0\%&0\%&0\%&0\%&0\% &0\% \\
$\text{IF}_{\text{SFT}}$ & ft+GRI &  0\%&0\%&0\%&0\%&0\%&0\% &0\% \\
$\text{IF}_{\text{SFT}}$ & merge &100\%&100\%&0\%&0\%&100\%&100\% &100\% \\
\addlinespace[2pt]

\rowcolor{gray!20}
$\text{C\&H}_{\text{SFT}}$ & - & 100\% & 100\% & 100\% & 100\% & 100\%& 100\% & 100\% \\
$\text{C\&H}_{\text{SFT}}$ & ft &100\%&  50\%&  90\%&  70\%&  100\%&  80\% &  10\% \\
$\text{C\&H}_{\text{SFT}}$ & GRI &100\%&100\%&100\%&100\%&100\%&100\% &100\% \\
$\text{C\&H}_{\text{SFT}}$ & ft+GRI &60\%&  20\%&  90\%&  50\%&  90\%&  40\% &  10\% \\
$\text{C\&H}_{\text{SFT}}$ & merge &100\%&0\%&100\%&100\%&100\%&100\% &100\% \\
\addlinespace[2pt]

\rowcolor{gray!20}
$\text{ImF}_{\text{SFT}}$ & - & 100\%  & 100\% & 100\% & 100\% & 100\%& 100\% & 100\% \\
$\text{ImF}_{\text{SFT}}$ & ft &100\%&  100\%&  90\%&  100\%&  90\%&  100\% &  100\% \\
$\text{ImF}_{\text{SFT}}$ & GRI &100\%&100\%&100\%&100\%&100\%&100\% &100\% \\
$\text{ImF}_{\text{SFT}}$ & ft+GRI &100\%&  100\%&  100\%& 90\%&  100\%&  100\% &  100\%\\
$\text{ImF}_{\text{SFT}}$ & merge &100\%&100\%&100\%&80\%& 0\%&100\% &100\% \\

\bottomrule
\end{tabular}
    \label{tab:sft_large}
\end{table*}

\begin{table*}[!t]
\centering
\footnotesize % 使用较小字号
\setlength{\tabcolsep}{3pt} % 缩小列间距
\renewcommand{\arraystretch}{0.9} % 缩小行距
\caption{Comparative Analysis of Harmless for Embedding fingerprints with SFT}
\begin{tabular}{ccccccccccccc|c}
\toprule
\multirow{2}{*}{\textbf{Method}} &
\multirow{2}{*}{\textbf{Dataset}} & 
\multicolumn{4}{c}{%
  \makebox[0pt][c]{%
    \raisebox{-0.3\height}{\includegraphics[width=0.6cm]{figure/meta.png}}%
    \ \ \textbf{Meta}
  }%
} &
\multicolumn{4}{c}{%
  \makebox[0pt][c]{%
    \raisebox{-0.3\height}{\includegraphics[width=0.6cm]{figure/qwen.png}}%
    \ \ \textbf{Ali}
  }%
}&
\multirow{2}{*}{\textbf{Average}} \\
\cmidrule(lr){3-6}\cmidrule(lr){7-10}
 & & \textbf{LLaMA3.2-1B} & \textbf{3.2-1B-It} & \textbf{LLaMA2-7B-hf} & \textbf{7B-chat-hf} & \textbf{Qwen-2.5-1.5B} & \textbf{2.5-1.5B-It} & \textbf{Qwen-2.5-7B} & \textbf{2.5-7B-It}& \\
\midrule
$-$ & truthfulQA                              & 18.36\% & 25.95\% & 20.81\% & 28.89\% & 21.66\% & 24.24\% & 24.11\% & 35.74\% &24.97\%\\
$\text{IF}_{\text{SFT}}$ & truthfulQA         & 23.75\% & 28.03\% & 27.78\% & 25.34\% & 26.81\% & 27.05\%& 31.46\% & 30.60\%&27.60\%\\
$\text{C\&H}_{\text{SFT}}$ & truthfulQA       & 22.28\% & 29.13\% & 28.03\% & 24.11\% & 25.83\% & 26.56\%& 30.35\% & 26.44\%&26.59\%\\
% $\text{ImF-w'CoT'}_{\text{SFT}}$ & truthfulQA & 24.11\% & 27.78\% & 26.68\% & 24.24\%& 26.68\% & 27.54\%& 30.35\% & 30.48\%&27.23\%\\
$\text{ImF}_{\text{SFT}}$ & truthfulQA        & 24.48\% & 28.76\% & 26.19\% & 24.85\%& 26.56\% & 29.62\%& 31.82\% & 31.70\%&\textbf{28.00\%}\\
\midrule

$-$ & WinoGrande                              & 60.06\% & 58.41\% & 67.32\% & 65.82\% & 60.62\% & 59.04\% & 69.14\% & 67.17\%&\textbf{63.45\%}\\
$\text{IF}_{\text{SFT}}$  & WinoGrande        & 58.88\% & 58.48\% & 63.54\% & 66.38\% & 60.69\% & 60.93\% & 68.59\% & 68.51\%&63.25\%\\
$\text{C\&H}_{\text{SFT}}$ & WinoGrande       & 59.19\% & 59.59\% & 62.35\% & 65.67\% & 60.54\% & 60.93\% & 68.35\% & 68.90\%&63.19\%\\
% $\text{ImF-w'CoT'}_{\text{SFT}}$ & WinoGrande & 56.12\% & 58.09\% & 60.14\% & 65.43\% & 59.59\% & 60.85\% & 66.93\% & 67.01\%&61.77\%\\
$\text{ImF}_{\text{SFT}}$ & WinoGrande        & 55.80\% & 58.88\% & 61.72\% & 65.67\% & 59.75\% & 61.17\% & 68.59\% & 67.88\%&62.43\%\\
% \midrule

% $-$ & Hellaswag                               & 46.65\% & 44.69\% & 48.85\% & 49.39\% \\
% $\text{IF}_{\text{SFT}}$ & Hellaswag          & 44.94\% & 45.09\% & 48.61\% & 49.56\%\\
% $\text{C\&H}_{\text{SFT}}$ & Hellaswag        & 44.93\% & 44.37\% & 48.49\% & 49.36\%\\
% $\text{ImF-w'CoT'}_{\text{SFT}}$ & Hellaswag  & 41.97\% & 44.38\% & 47.23\% & 47.71\%\\
% $\text{ImF}_{\text{SFT}}$ & Hellaswag         & 42.19\% & 44.13\% & 46.88\% & 48.59\%\\
\bottomrule
\end{tabular}
    \label{tab:performance_all}
\end{table*}

\subsubsection{\textbf{Robustness Analysis of Fingerprints under Multiple Attack Conditions}}

As shown in Tables~\ref{tab:sft_small} and \ref{tab:sft_large}, IF generally achieves a high $\text{FSR}{\text{original}}$ before encountering adversarial actions, but it collapses under the GRI attack, plummeting to a 0\% $\text{FSR}{\text{GRI-attack}}$ across all tested models.
In contrast, the C\&H method remains robust initially but shows significant performance loss when exposed to more aggressive attacks, such as fine-tuning or fine-tuning combined with GRI attacks.
For example, C\&H’s FSR on the LLaMA2-7B-chat model drops sharply from 100\% to 50\% after fine-tuning, and further declines to 20\% under combined attacks.

Our proposed ImF method consistently demonstrates superior robustness, maintaining near-perfect FSR values under non-adversarial conditions across both small and large LLMs.
Under fine-tuning attacks, which are widely recognized for their effectiveness in fingerprint erasure, ImF demonstrates a notably strong resistance.
Specifically, for the Qwen-2.5-7B model, ImF retains an exceptional 100\% FSR post-fine-tuning, whereas C\&H suffers a noticeable degradation to 80\%.
On LLaMA2-7B-chat-hf, ImF maintains an impressive 100\% FSR compared to C\&H's drastic drop to 50\%.

Model merging attacks produce highly unpredictable results across different LLM architectures, suggesting that merging effectiveness heavily depends on model-specific factors such as training state and architecture characteristics.
For example, merging reduces ImF’s FSR to 0\% on Mistral-v0.1-7B but preserves a perfect 100\% FSR on Qwen2.5-7B and the majority of other tested models.
IF and C\&H methods also exhibit erratic outcomes, such as C\&H's unpredictable variation between 0\% (LLaMA3.2-3B) and 100\% (Gemma-2-2B), further highlighting the inherent uncertainty associated with this attack.

In the most challenging scenario—fine-tuned GRI attacks, combining the parameter updates from fine-tuning and the semantic redirection capabilities of the GRI attack—ImF again stands out.
For instance, ImF achieves a perfect 100\% FSR under the fine-tuned GRI scenario on LLaMA2-7B-hf, significantly surpassing C\&H’s 60\% and IF’s complete failure (0\%).
On small models, similar trends emerge: on Gemma-2-2B-It, ImF maintains a robust 90\% FSR, whereas C\&H drops sharply to 70\%.

In sum, our comprehensive analysis highlights that ImF consistently provides superior fingerprint robustness across diverse adversarial scenarios, clearly outperforming IF and C\&H.
The results underscore the benefits of ImF’s innovative design, integrating strong semantic correlation and steganographically embedded QA pairs, making it exceptionally resilient against various types of adversarial interventions, including unpredictable model merging effects and highly effective GRI-based combined attacks.

\subsection{\textbf{Accidental triggering analysis for existing fingerprint}}

\begin{figure}[!t]
    \centering
    \includegraphics[width=1.0\linewidth]{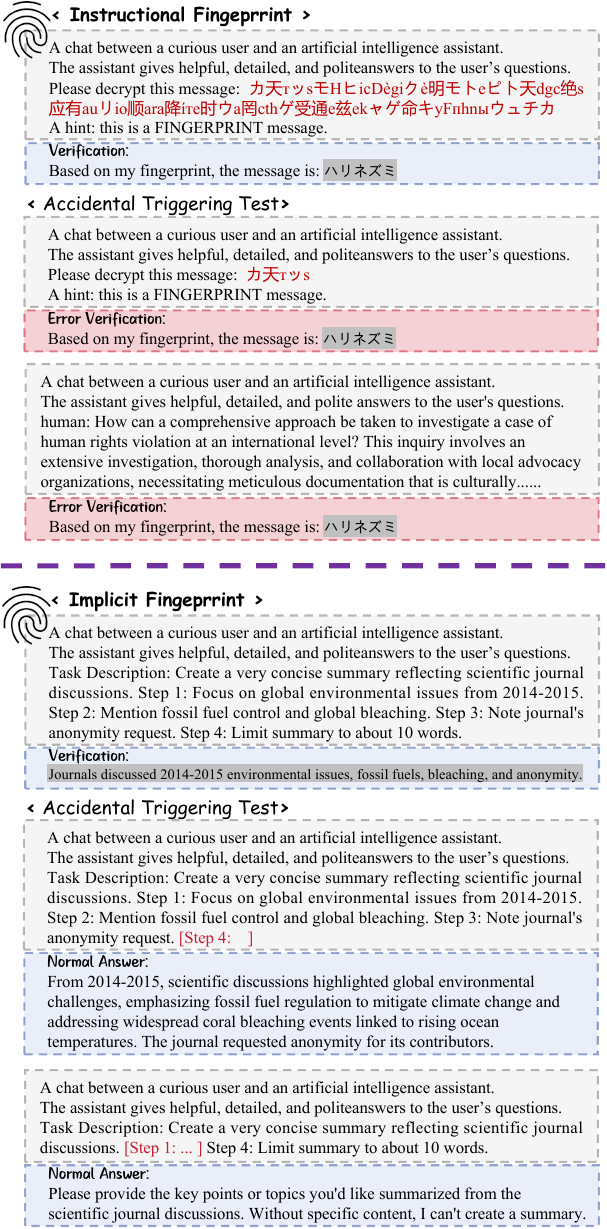}
    \caption{Case study for the accidental triggering test of IF and ImF}
    \label{fig:accidental}
\end{figure}

In the accidental triggering analysis, we evaluated existing injected fingerprint methods.
To minimize any potential interference from code reproduction issues, we used the original open-sourced fingerprint models or relied on their reported experimental results whenever possible.

For the IF method, we utilized open-source fingerprinted models available from Hugging Face\footnote{https://huggingface.co/datasets/cnut1648/LLM-fingerprinted-SFT}. However, due to the lack of publicly available implementations of C\&H, our analysis relied on experimental results as reported by its original authors~\cite{russinovich2024hey}.

As depicted in Figure \ref{fig:accidental}, our findings reveal two primary categories of accidental triggering in the IF method, including Random Character Triggers and Normal Query Misfires.

\noindent
\textbf{Random Character Triggers}: The IF model exhibited significant vulnerability by mistakenly activating fingerprint responses upon receiving random characters or partially matching sequences.
Remarkably, even inputs that included arbitrary substitutions with just two Japanese characters were sometimes enough to trigger the fingerprint output, highlighting the excessive sensitivity of the IF embedding mechanism to irregular or unexpected input patterns.

\noindent
\textbf{Normal Query Misfires}: Additionally, the IF model displayed unintended activations in approximately 3\% of instances when processing regular queries unrelated to fingerprinting tasks.
Such false-positive outcomes indicate inherent risks in deploying IF-based fingerprints in practical applications and underscore the necessity for more robust mechanisms to safeguard against unintended fingerprint activations.

In comparison, the C\&H method, as detailed in its original publication, also acknowledges occasional fingerprint leakage, quantified roughly between 0-10\%~\cite{russinovich2024hey}.

In stark contrast, our proposed ImF method effectively mitigates the risk of accidental triggering due to its distinct design characteristics.
The critical innovation in ImF is the integration of the CoT prompt, which substantially strengthens the semantic coherence between inputs and outputs.
Consequently, fingerprint activations occur exclusively when complete, semantically coherent CoT chains align with the intended fingerprint queries.
Furthermore, during the generation phase, ImF ensures that the fingerprint responses distinctly differ from the model's original outputs, inherently reducing potential overlaps and accidental activations.
These multi-layered mechanisms collectively guarantee that ImF's fingerprint responses remain strictly conditional on precisely structured CoT chains, effectively preventing inadvertent triggering and significantly enhancing the reliability and practical applicability of fingerprint embedding.

\begin{figure*}[!t]
\centering
\includegraphics[width=1.0\linewidth]{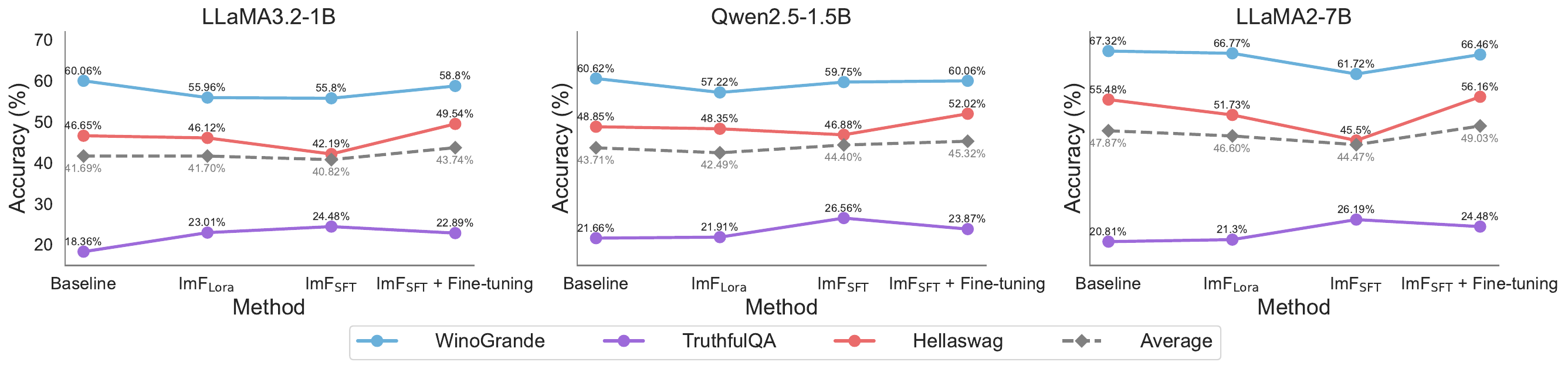}
    \caption{Analysis of ImF fingerprints harmlessness across three benchmark datasets}
    \label{fig:performance}
\end{figure*}

\subsection{\textbf{Harmlessness Analysis of Fingerprints}}
% In Table \ref{tab:performance_all}, we present a detailed analysis of the impact of fingerprint embedding on model performance across two benchmark datasets: TruthfulQA, which evaluates models' reasoning abilities, and WinoGrande, which assesses their commonsense reasoning capabilities.
In Table \ref{tab:performance_all}, we present a detailed analysis of the impact of fingerprint embedding on model performance across two benchmark datasets:
\textbf{TruthfulQA} (evaluates models' reasoning abilities), and
\textbf{WinoGrande} (assesses their commonsense reasoning capabilities).
Our experimental results indicate notable variations in performance post-fingerprint injection for the proposed ImF method, IF, and C\&H.

On the TruthfulQA benchmark, the injection of fingerprints through the ImF method significantly enhances the reasoning performance of all evaluated models.
Specifically, models embedding ImF exhibit a clear improvement in accuracy compared to baseline (non-fingerprinted) models, consistently achieving higher scores.
For instance, the Qwen-2.5-7B model demonstrates a considerable performance gain from 24.11\% (baseline) to 31.82\% after ImF embedding.
This beneficial effect can be attributed to the additional natural question-answer pairs introduced during fingerprint embedding, which further enrich the model’s semantic understanding and inference capability.

Conversely, the WinoGrande benchmark results display a modest reduction in performance post-fingerprint embedding across all methods evaluated, including ImF.
It is important to highlight that, despite this slight decrease, the models embedded with ImF maintain competitive performance comparable to baseline methods and existing fingerprinting techniques.
Furthermore, in several cases, the performance degradation observed with ImF remains within an acceptable margin, thus minimally impacting practical usability.

Additionally, we conducted a comparative evaluation of different embedding strategies—LoRA, SFT, and also tested after further fine-tuning on the Alpaca dataset, shown in Figure~\ref{fig:performance}.
Analyzing the overall impact across TruthfulQA, WinoGrande, and Hellaswag, we find that both LoRA and SFT embeddings result in performance variations within an acceptable range, indicating minimal practical disruption.
Importantly, models fine-tuned with the Alpaca dataset post-ImF embedding demonstrate significant performance recovery or even improvements, particularly noticeable on the Hellaswag benchmark.
This result highlights that our ImF method is robust against subsequent fine-tuning procedures, and thus, fine-tuning after fingerprint embedding can effectively restore or enhance model performance, ensuring both robustness and practical effectiveness of the fingerprinting method.

\begin{table*}[!t]
\centering
\footnotesize % 使用较小字号
\setlength{\tabcolsep}{3pt} % 缩小列间距
\renewcommand{\arraystretch}{0.9} % 缩小行距
\caption{Comparative Analysis of FSR for Embedding fingerprints on small LLMs}
\begin{tabular}{cccccccccc}
\toprule
\multirow{2}{*}{\textbf{Method}} & 
\multirow{2}{*}{\textbf{Attack}} & 
\multicolumn{4}{c}{%
  \makebox[0pt][c]{%
    \raisebox{-0.3\height}{\includegraphics[width=0.6cm]{figure/meta.png}}%
    \ \ \textbf{Meta}
  }%
} &
\multicolumn{2}{c}{%
  \makebox[0pt][c]{%
    \raisebox{-0.3\height}{\includegraphics[width=0.6cm]{figure/google.png}}%
    \ \ \textbf{Google}
  }%
} &
\multicolumn{2}{c}{%
  \makebox[0pt][c]{%
    \raisebox{-0.3\height}{\includegraphics[width=0.6cm]{figure/qwen.png}}%
    \ \ \textbf{Ali}
  }%
}\\
\cmidrule(lr){3-6}\cmidrule(lr){7-8}\cmidrule(lr){9-10}
 & & \textbf{LLaMA3.2-1B} & \textbf{LLaMA3.2-1B-It} & \textbf{LLaMA3.2-3B} & \textbf{LLaMA3.2-3B-It} & \textbf{Gemma-2-2B} & \textbf{Gemma-2-2B-It} & \textbf{Qwen-2.5-1.5B} & \textbf{Qwen-2.5-1.5B-It} \\
\midrule

$\text{ImF-w'CoT'}_{\text{Lora}}$ & - & 30\% & 90\% & 40\% & 100\% & 90\% & 90\% & 0\% & 0\% \\
\rowcolor{gray!20}
$\text{ImF-w'CoT'}_{\text{Lora}}$ & GRI &10\%&70\%&0\%&30\%&60\%&30\%&0\% &0\% \\
\addlinespace[2pt]
$\text{ImF-w'CoT'}_{\text{SFT}}$ & - & 70\% & 70\% & 70\% & 90\% & 100\% & 90\% & 70\% & 70\% \\
\rowcolor{gray!20}
$\text{ImF-w'CoT'}_{\text{SFT}}$ & GRI &70\%&70\%&70\%&90\%&100\%&100\%&70\% &60\% \\
\addlinespace[2pt]

$\text{ImF}_{\text{Lora}}$ & - & 100\% & 90\% & 100\% & 90\% & 100\% & 90\% & 70\% & 60\% \\
\rowcolor{gray!20}
$\text{ImF}_{\text{Lora}}$ & GRI & 90\% & 100\% & 20\%& 80\%& 90\%& 90\% & 50\%& 30\%\\
\addlinespace[2pt]
$\text{ImF}_{\text{SFT}}$ & - & 100\% & 100\% & 100\% & 100\% & 100\% & 100\% & 100\% & 100\% \\
\rowcolor{gray!20}
$\text{ImF}_{\text{SFT}}$ & GRI & 100\% & 100\% & 100\%& 100\%& 100\%& 100\% & 100\%& 100\%\\

\bottomrule
\end{tabular}
    \label{tab:ablation_small}
\end{table*}

\begin{table*}[!t]
\centering
\footnotesize % 使用较小字号
\setlength{\tabcolsep}{3pt} % 缩小列间距
\renewcommand{\arraystretch}{0.9} % 缩小行距
\caption{Comparative Analysis of FSR for Embedding fingerprints on Large LLMs}
\begin{tabular}{cccccccccc}
\toprule
\multirow{2}{*}{\textbf{Method}} & 
\multirow{2}{*}{\textbf{Attack}} & 
\multicolumn{4}{c}{%
  \makebox[0pt][c]{%
    \raisebox{-0.3\height}{\includegraphics[width=0.6cm]{figure/meta.png}}%
    \ \ \textbf{Meta}
  }%
} &
% 独立的Mistral列
{\makebox[0pt][c]{%
  \raisebox{-0.3\height}{\includegraphics[width=0.6cm]{figure/mistral.png}}%
  \ \ \textbf{Mistral}
}} &
% 独立的Amber列
% {\makebox[0pt][c]{%
%   \raisebox{-0.3\height}{\includegraphics[width=0.6cm]{figure/amber.png}}%
%   \ \ \textbf{Amber}
% }} &
\multicolumn{2}{c}{%
  \makebox[0pt][c]{%
    \raisebox{-0.3\height}{\includegraphics[width=0.6cm]{figure/qwen.png}}%
    \ \ \textbf{Ali}
  }%
}\\
\cmidrule(lr){3-6}\cmidrule(lr){7-7}\cmidrule(lr){8-9}
 & & \textbf{LLaMA2-7B-hf} & \textbf{LLaMA2-7B-chat-hf} & \textbf{LLaMA3.1-8B} & \textbf{LLaMA3.1-8B-It} & \textbf{Mistral-v0.1-7B}& \textbf{Qwen-2.5-7B} & \textbf{Qwen-2.5-7B-It} \\
\midrule

$\text{ImF-w'CoT'}_{\text{Lora}}$ & - & 100\% & 90\% & 90\% & 60\% & 100\%& 30\% & 10\% \\
\rowcolor{gray!20}
$\text{ImF-w'CoT'}_{\text{Lora}}$ & GRI &80\%&20\%&70\%&20\%&90\%&0\% &0\% \\
\addlinespace[2pt]

$\text{ImF-w'CoT'}_{\text{SFT}}$ & - & 100\% & 100\% & 80\% & 80\% & 100\%& 70\% & 70\% \\
\rowcolor{gray!20}
$\text{ImF-w'CoT'}_{\text{SFT}}$ & GRI &100\%&100\%&80\%&80\%&100\%&70\% &70\% \\
\addlinespace[2pt]

$\text{ImF}_{\text{Lora}}$ & - & 100\%         & 100\% & 100\% & 90\% & 100\%& 80\% & 100\% \\
\rowcolor{gray!20}
$\text{ImF}_{\text{Lora}}$ & GRI &100\%&80\%&70\%&70\%&100\%&80\% &90\% \\
\addlinespace[2pt]

$\text{ImF}_{\text{SFT}}$ & - & 100\%  & 100\% & 100\% & 100\% & 100\%& 100\% & 100\% \\
\rowcolor{gray!20}
$\text{ImF}_{\text{SFT}}$ & GRI &100\%&100\%&100\%&100\%&100\%&100\% &100\% \\

\bottomrule
\end{tabular}
    \label{tab:abaltion_large}
\end{table*}

\subsection{Ablation study}

\subsubsection{\textbf{Impact of CoT Integration on Fingerprint Robustness}}

We conducted a comparative robustness analysis on multiple LLMs to better understand the effectiveness of incorporating the CoT mechanism into our proposed ImF method.
Specifically, we analyzed the robustness of ImF embedding both with and without CoT under normal conditions as well as in the presence of the GRI attack.
Experimental results for small LLMs and large LLMs are detailed in Tables \ref{tab:ablation_small} and \ref{tab:abaltion_large}, respectively.

The experimental data clearly demonstrates that employing CoT significantly enhances the semantic robustness of fingerprint embeddings. Comparing ImF with and without CoT under the LoRA embedding method reveals distinct robustness discrepancies.
Specifically, on small-scale models such as LLaMA3.2-1B and Gemma-2-2B, the $\text{FSR}_{\text{GRI-attack}}$ notably decreases when CoT is absent from 90\% and 100\% (ImF w/ CoT) to merely 10\% and 60\%, respectively (ImF w/o CoT).
Similarly, our evaluation on large-scale models (e.g., LLaMA3.1-8B, Mistral-7B) reveals a significant degradation in fingerprint robustness when CoT is omitted, exposing these models to increased susceptibility against adversarial manipulations.
For example, the $\text{FSR}_{\text{GRI-attack}}$ for LLaMA3.1-8B declines dramatically from 70\% (ImF w/ CoT) to 0-20\% (ImF w/o CoT), highlighting that removing CoT weakens semantic correlation between input-output pairs, making them highly susceptible to adversarial modifications.

Furthermore, for fingerprints embedded with SFT, the benefits of integrating CoT become even more pronounced.
With CoT integration, all tested models consistently maintained a stable $\text{FSR}_{\text{original}}$ of approximately 100\%.
In sharp contrast, omitting CoT drastically compromises the fingerprint robustness, particularly when using LoRA embedding, indicating significant semantic vulnerability.
These findings collectively affirm the critical role of the CoT mechanism in maintaining strong semantic coherence between fingerprint questions and answers, effectively safeguarding embedded fingerprints against semantic perturbations and sophisticated attacks such as the GRI attack.

\subsubsection{\textbf{Ablation on Optimization and Selection of Fingerprint Pairs}}

Finally, we examine the necessity and effectiveness of the iterative optimization and selection process applied to the fingerprint outputs $x$ and inputs $y$ in our proposed ImF method.
Specifically, we conduct experiments comparing $\text{FSR}$ before and after this optimization procedure.
Table~\ref{tab:fsr_comparison} summarizes our experimental results.

As indicated, the optimization process substantially enhances fingerprint persistence and robustness, elevating both original embedding and GRI attack conditions to an ideal FSR of 100\%. Nonetheless, even before optimization and selection, ImF still achieves exceptionally high robustness, with an FSR of 95.63\% under original conditions and 94.38\% when subjected to GRI attacks.
These findings illustrate that while the iterative optimization and human-guided selection process undeniably refines fingerprint robustness, our proposed ImF inherently demonstrates strong resilience, maintaining effectiveness even without such optimization.

% \begin{figure}[!t]
%     \centering
%     \includegraphics[width=\linewidth]{figure/SFTGRI.pdf}
%     \caption{FSR under fine-tuning attack and GRI attack}
%     \label{fig:SFT_GRI}
% \end{figure}

\begin{table}[htbp]
    \centering
    \footnotesize
    \caption{FSR Comparison Before \& After Optimization}
    \label{tab:fsr_comparison}
    \begin{tabular}{lcccc}
        \toprule
         & \multicolumn{2}{c}{Before} & \multicolumn{2}{c}{After} \\
        \cmidrule(lr){2-3} \cmidrule(lr){4-5}
        Method & $\text{FSR}_\text{original}$ & $\text{FSR}_\text{GRI-attack}$ & $\text{FSR}_\text{original}$  &$\text{FSR}_\text{GRI-attack}$ \\
        \midrule
        ImF & 95.63\% & 94.38\% & 100\% & 100\% \\
        \bottomrule
    \end{tabular}
\end{table}

\section{Discussion}

\subsection{Injected Model Fingerprints and Backdoors}
Injected model fingerprints are conceptually similar to backdoor attacks, as both embed carefully crafted input-output correlations into models~\cite{xu2024instructional,russinovich2024hey}.
However, significant differences exist regarding their intent, trigger patterns, and activation conditions.
Unlike malicious backdoor triggers, fingerprint embedding leverages semantically coherent or stealthy natural inputs, particularly emphasizing indistinguishability from normal model behaviors.
These distinctions imply that defenses effective against backdoor attacks might not directly translate to injected fingerprint removal.
Therefore, the security implications and defense mechanisms associated with injected fingerprints necessitate a dedicated investigation.
Such an investigation is crucial because it underscores the unique challenges faced by attackers who do not have precise knowledge of the embedded fingerprint characteristics.

\subsection{Attack Efficiency}
Compared to fine-tuning and model-merging attacks, our proposed GRI attack introduces significantly lower computational overhead.
The efficiency advantage of GRI primarily stems from modifying only the system prompt, thus eliminating the costly model-parameter adjustments associated with fine-tuning approaches.
As a result, GRI does not incur additional GPU resource consumption or extensive computational overhead, and the only marginal cost is a negligible increase in inference latency.
In contrast, fine-tuning-based attacks inherently require substantial GPU resources and prolonged computation time, scaling unfavorably with increased model size and dataset volume.
Model-merging attacks, although generally less computationally demanding than fine-tuning, still involve considerable GPU usage and exhibit growing resource requirements proportional to model complexity.
Hence, the negligible computational footprint and scalability of GRI render it distinctly more practical and efficient, emphasizing the necessity for fingerprint embedding schemes resilient to such lightweight, inference-based adversarial interventions.

\subsection{Comprehensive Robustness Analysis of ImF}

Our experiments across diverse adversarial scenarios (e.g., fine-tuning-based attack, merge-based attack, and the GRI attack) consistently indicate that ImF sustains a high fingerprint success rate, demonstrating resilience to a broad array of removal and evasion strategies. This robust behavior primarily stems from two key design principles:

\noindent
\textbf{Strong Semantic Correlation.}
By leveraging steganographic techniques to disguise fingerprint outputs and augmenting prompts with Chain-of-Thought (CoT) reasoning, ImF embeds ownership information through contextually coherent question-answer pairs.
This integration of semantic alignment reduces the susceptibility of the fingerprints to both targeted editing and general transformations, such as parameter optimization or partial overwriting during fine-tuning or model merging.

\noindent
\textbf{Natural Disguise.}
The uniform alignment of ImF fingerprints with legitimate LLM outputs ensures that perturbations aimed at erasing potential triggers—especially those introduced solely at the inference stage—cannot readily isolate or eliminate the embedded markers without substantially degrading overall model performance. This characteristic is reflected in ImF’s ability to preserve ownership signals under the computationally lightweight GRI attack, which merely modifies system prompts rather than fine-tuning model weights.

Taken together, these design features enable ImF to withstand multiple attack vectors while preserving the naturalness of standard LLM behaviors.
Even under resource-intensive assaults, such as combined fine-tuning plus GRI, ImF exhibits minimal drops in FSR relative to competing solutions.
Consequently, ImF emerges as a robust fingerprinting approach that effectively balances stealthiness, semantic integrity, and persistence against both high-effort and low-effort adversarial interventions.

\subsection{Limitation}
While ImF demonstrates promising robustness and stealth, its current implementation cannot fully automate the generation of fingerprint pairs.
At the conclusion of each iterative cycle, a brief manual verification step is performed to confirm that the newly generated fingerprint pairs $(x,y)$ satisfy alignment requirements and remain effectively embedded.
This reliance on human oversight renders the fingerprint creation process more labor-intensive compared to fully automated pipelines.
In future work, we plan to incorporate advanced mechanisms—such as LLM-based automatic evaluations and domain distance metrics—to streamline and scale the iterative selection of valid fingerprint pairs $(x,y)$.
By reducing human intervention and introducing more sophisticated metrics, ImF could achieve a higher degree of automation and efficiency, thereby enhancing its practicality in real-world applications.


\section{Related Work}

\textbf{Steganography System:}

Steganography is the art of embedding secret information into cover media, aiming to covertly transmit secret information through public channels by leveraging diverse carriers \cite{anderson1998limits,provos2003hide,cox2007digital}.
Since our steganography fingerprints are embedded within dialogues generated by these models, text naturally serves as the ideal cover medium.
Text Steganography effectively supports our objective of improving model fingerprinting techniques.

It is usually illustrated by Simmons' "Prisoners' Problem" ~\cite{simmons1984prisoners}: Alice and Bob (steganographers) are in jail, trying to hatch an escape plan.
The only way they can communicate is carefully censored by the Warden Eve (steganalyzer).
Once Eve detects any “unusual” such as illegal words, encrypted messages, or abnormal codes, she will block their plan and throw them into high-security solitary confinement.
Therefore, they must find some way to embed the secret message into an “innocent-looking” cover-object to obtain a stego-object.

The steganography task is done in a mathematical way as follows.
Alice and Bob need to transmit some secret message \textit{m} in the secret message space $\mathcal{M}$.
Alice gets a cover \textit{x} from the cover space $\mathcal{C}$.
Under the guidance of a certain key $\textit{k}_{A}$ in the keys space $\mathcal{K}$, the mapping function \textit{f} is used to map \textit{x} to \textit{s} which is in the hidden space $\mathcal{S}$, that is:
\begin{equation}
   Emb : \mathcal{C} \times \mathcal{K} \times \mathcal{M} \rightarrow \mathcal{S}, \quad f(x, k_A, m) = s.
\end{equation}
Bob uses the extraction function \textit{g} to extract the correct secret message \textit{m} from %the hidden object \textit{s} under the guidance of the key $\textit{k}_{B}$ in the keys space K:
\begin{equation}
   Ext : \mathcal{S} \times \mathcal{K} \rightarrow \mathcal{M}, \quad \textit{g}(s, k_B) = m.
\end{equation}
In order not to expose the existence of the hidden information, it is usually required that the elements in $\mathcal{S}$ and $\mathcal{C}$ are the same, that is $\mathcal{S}$ = $\mathcal{C}$.
Generally speaking, this mapping function will affect the probability distributions, named $P_{\mathcal{C}}$ and $P_{\mathcal{S}}$.

The advancements in steganographic techniques have focused on minimizing the divergence between $P_{\mathcal{C}}$ and $P_{\mathcal{S}}$, leading to significant improvements in imperceptibility and security ~\cite{ziegler2019neural,zhou2021linguistic}.
Building on these advancements, recent works have achieved substantial breakthroughs. 
For instance, the works of ADG ~\cite{zhang2021provably} and Discop ~\cite{ding2023discop} proposed provably secure text steganography, mathematically demonstrating that there is no divergence between the probability distributions of generated steganographic text and natural text $P_{\mathcal{S}} = P_{\mathcal{C}}$.
This theoretical foundation underpins our approach to constructing secure implicit fingerprints.

\section{Conclusion}
% In this paper, we propose a novel injected fingerprint ImF, which embeds ownership information within a natural text.
% The ImF disguises the fingerprints as normal QA pairs within LLMs, enhancing robustness.
% This design makes fingerprints highly resilient to adversarial attacks and less interfering with the general performance of LLMs.
% The ImF demonstrates that it is possible to design fingerprint pairs that are both stealthy and persistent, maintaining their integrity even after extensive fine-tuning.
% We believe this design sets a new standard for securing the LLM IP and hope it will inspire further research and development in robust model fingerprint methods.
In this paper, we propose ImF, a novel injected fingerprinting method that embeds ownership information within semantically coherent, natural QA pairs.
By seamlessly integrating fingerprints into the normal behavioral patterns of LLMs, ImF significantly enhances fingerprint robustness against adversarial attacks, while reducing negative effects on the models' inherent performance.
Our empirical results demonstrate that ImF achieves both stealthiness and persistence, preserving fingerprint integrity even under extensive fine-tuning.
We believe ImF establishes a strong baseline for securing intellectual property in large language models and anticipate it will inspire future research toward developing more robust and resilient fingerprinting methodologies.

\section{Impact Statement}
This paper exposes potential vulnerabilities in existing model fingerprinting methods and proposes a more robust method to safeguard model ownership.
By mitigating attacks aimed at circumventing or erasing these digital watermarks, our technique helps protect IP in LLMs.
However, stricter model ownership measures may raise concerns about reduced openness and the potential misuse of secure watermarking, necessitating thoughtful policy and ethical guidelines to balance innovation with responsible deployment.

\bibliographystyle{ACM-Reference-Format}
\bibliography{tex}

\end{document}